\documentclass{bmvc2k}

\title{CounTR: Transformer-based Generalised Visual Counting}

\addauthor{Chang Liu}{liuchang666@sjtu.edu.cn}{1}
\addauthor{Yujie Zhong}{jaszhong@hotmail.com}{2}
\addauthor{Andrew Zisserman}{az@robots.ox.ac.uk}{3}
\addauthor{Weidi Xie}{weidi@sjtu.edu.cn}{1}

\addinstitution{
Coop. Medianet Innovation Center, \\ Shanghai Jiao Tong University, China
}
\addinstitution{
 Meituan Inc., China
}
\addinstitution{
Visual Geometry Group~(VGG), \\ University of Oxford, UK}

\runninghead{Liu, et. al.}{Transformer-based Generalised Visual Counting}

\newcommand{\weidi}[1]{{\textcolor{magenta}{[Weidi: #1]}}}

\usepackage[svgnames]{xcolor}
\usepackage{caption}
\definecolor{bmvc_blue}{RGB}{0,0,102} 
\usepackage{bm}
\usepackage{bbding}
\usepackage{comment}
\usepackage{amssymb}
\usepackage{multirow}
\usepackage{booktabs}
\usepackage{subfigure}
\usepackage{todonotes}
\usepackage{graphicx}
\usepackage{caption}
\usepackage{appendix}

\begin{document}

\maketitle

\begin{abstract}

In this paper, 
we consider the problem of generalised visual object counting, 
with the goal of developing a computational model for counting the number of objects from {\em arbitrary} semantic categories, using {\em arbitrary} number of ``exemplars'', {\em i.e.}~zero-shot or few-shot counting.
To this end, we make the following four contributions: 
(1) We introduce a novel transformer-based architecture for generalised visual object counting, termed as Counting TRansformer~(\textbf{CounTR}), 
which explicitly captures the similarity between image patches or with given ``exemplars'' using the attention mechanism;
(2) We adopt a two-stage training regime, that first pre-trains the model with self-supervised learning, and followed by supervised fine-tuning;
(3) We propose a simple, scalable pipeline for synthesizing training images with a large number of instances or that from different semantic categories, 
explicitly forcing the model to make use of the given ``exemplars'';
(4) We conduct thorough ablation studies on the large-scale counting benchmark,
{\em e.g.}~FSC-147, and demonstrate state-of-the-art performance on both zero and few-shot settings.
Project page: \url{https://verg-avesta.github.io/CounTR_Webpage/}.
\end{abstract}

\section{Introduction}

Despite all the exceptional abilities, 
the human visual system is particularly weak in counting objects in the image. 
In fact, given a visual scene with a collection of objects, 
one can only make a rapid, accurate, and confident judgment if the number of items is below five, with an ability termed as subitizing~\cite{Kaufman49}.
While for scenes with an increasing number of objects,
the accuracy and confidence of the judgments tend to decrease dramatically. 
Until at some point, counting can only be accomplished by calculating estimates 
or enumerating the instances, which incurs low accuracy or tremendous time cost.

In this paper, our goal is to develop a {\bf generalised visual object counting} system, that augments humans' ability for recognising the number of objects in a visual scene. Specifically, generalised visual object counting refers to the problem of identifying the number of the salient objects of {\em arbitrary} semantic class in an image ({\em i.e.}~open-world visual object counting)
with {\em arbitrary} number of instance ``exemplars'' provided by the end user, to refer to the particular objects to be counted, {\em i.e.}~from zero-shot to few-shot object counting.
To this end, we propose a novel architecture that transforms the input image (with the few-shot annotations if any) into a density map, and the final count can be obtained by simply summing over the density map.

Specifically, we take inspiration from Lu {\em et al.}~\cite{Lu18}
that self-similarity is a strong prior in visual object counting, 
and introduce a transformer-based architecture
where the self-similarity prior can be explicitly captured by the built-in attention mechanisms, both among the input image patches and with the few-shot annotations~(if any). 
We propose a two-stage training scheme, 
with the transformer-based image encoder being firstly pre-trained with self-supervision via masked image modeling~\cite{he2022masked}, 
followed by supervised fine-tuning for the task at hand.
We demonstrate that self-supervised pre-training can effectively learn the visual representation for counting, thus significantly improving the performance.
Additionally, to tackle the long-tailed challenge in existing generalised visual object counting datasets, where the majority of images only contain a small number of objects, we propose a simple, yet scalable pipeline for synthesizing training images with a large number of instances,
as a consequence, establishing reliable data sources for model training,
to condition the user-provided instance exemplars.


To summarise, in this paper, we make four contributions:
{\em First}, we introduce an architecture for generalised visual object counting based on transformer, termed as \textbf{CounTR}~(pronounced as counter). 
It exploits the attention mechanisms to explicitly capture the similarity between image patches, or with the few-shot instance ``exemplars'' provided by the end user;
{\em Second}, we adopt a two-stage training regime~(self-supervised pre-training, followed by supervised fine-tuning) and show its effectiveness for the task of visual counting;
{\em Third}, we propose a simple yet scalable pipeline for synthesizing training images with a large number of instances, 
and demonstrate that it can significantly improve the performance on images containing a large number of object instances;
{\em Fourth}, we conduct thorough ablation studies on the large-scale counting benchmark, {\em e.g.}~FSC-147~\cite{Ranjan21}, and demonstrate state-of-the-art performance on both zero-shot and few-shot settings, improving the previous best approach by over 18.3\% on the mean absolute error of the test set.

\vspace{-10pt}
\section{Related Work}
\paragraph{Visual object counting.}
In the literature, object counting approaches can generally be cast into two categories: detection-based counting~\cite{barinova2012detection,desai2011discriminative,hsieh2017drone} or regression-based counting~\cite{arteta2014interactive,arteta2016counting,cho1999neural,kong2006viewpoint,lempitsky2010learning,marana1997estimation,xie2018microscopy}. 
The former relies on a visual object detector that can localize object instances in an image.
This method, however, requires training individual detectors for different objects, 
and the detection problem remains challenging if only a small number of annotations are given.
The latter avoids solving the hard detection problem, 
instead, methods are designed to learn either a mapping from global image features to a scalar (number of objects), 
or a mapping from dense image features to a density map, 
achieving better results on counting overlapping instances. 
However, previous methods from both lines~(detection, regression) have only been able to count objects of one particular class~({\em e.g.}~cars, cells).\\[-0.8cm]

\paragraph{Class-agnostic object counting.}
Recently, class-agnostic few-shot counting~\cite{Lu18,Ranjan21,you2022iterative} has witnessed a rise in research interest in the community.
Unlike the class-specific models that could only count objects of specific classes like cars, cells, or people, class-agnostic counting aims to count the objects in an image based on a few given ``exemplar'' instances,
thus is also termed as few-shot counting.
Generally speaking, class-agnostic few-shot counting models need to mine the commonalities between the counts of different classes of objects during training. In~\cite{Lu18}, the authors propose a generic matching network~(GMN), which regresses the density map by computing the similarity between the CNN features from image and exemplar shots;
FamNet~\cite{Ranjan21} utilizes feature correlation for prediction and uses adaptation loss to update the model's parameters at test time;
SAFECount~\cite{you2022iterative} uses the support feature to enhance the query feature, making the extracted features more refined and then regresses to obtain density maps;
In a very recent work~\cite{Hobley22}, 
the authors exploit a pre-trained DINO~\cite{Mathilde21} model and a lightweight regression head to count without exemplars.
In this paper, we also use transformer-based architecture, 
however, train it from scratch, and augment it with the ability to count the objects with {\em any shot}.

\section{Methods}

In this paper, 
we consider the challenging problem of generalised visual object counting, where the goal is to count the salient objects of \textbf{arbitrary} semantic class in an image, {\em i.e.}~open-world visual object counting, with \textbf{arbitrary} number of ``exemplars'' provided by the end user,{\em i.e.}~from zero-shot to few-shot object counting.\\[-0.8cm]

\paragraph{Overview.}
Given a training set, 
{\em i.e.}~$\mathcal{D}_{\text{train}} = \{(\mathcal{X}_1, \mathcal{S}_1, {y}_1), \dots, (\mathcal{X}_N, \mathcal{S}_N, {y}_N)\}$, where $\mathcal{X}_i \in \mathbb{R}^{H \times W \times 3}$ denotes the input image, 
$\mathcal{S}_i = \{b_i\}^K$ denotes the box coordinates~($b_i^k \in \mathbb{R}^4$) for a total of $K \in \{0,1,2,3...\}$ given ``exemplars'', 
{\em i.e.}~zero-shot or few-shot counting, 
$y_i \in \mathbb{R}^{H \times W \times 1}$ refers to a binary spatial density map, with $1$'s at the objects' center location, 
indicating their existence, and $0$'s at other locations without the objects, 
the object count can thus be computed by spatially summing over the density map. Our goal here is thus to train a generalised visual object counter that can successfully operate on a test set, given zero or few exemplars, {\em i.e.}~$\mathcal{D}_{\text{test}} = \{(\mathcal{X}_{N+1}, \mathcal{S}_{N+1}), \dots, (\mathcal{X}_{M}, \mathcal{S}_{M})\}$.
\textbf{Note that}, the semantic categories for objects in training set~($\mathcal{C}_{\text{train}}$) and testing set~($\mathcal{C}_{\text{test}}$) are disjoint, {\em i.e.}~$\mathcal{C}_{\text{train}} \cap \mathcal{C}_{\text{test}} = \emptyset$.

To achieve this goal,
we introduce a novel transformer-based architecture, 
termed as Counting TRansformer~(\textbf{CounTR}).
Specifically, the attention mechanisms in transformer enable to explicitly compare visual features between any other spatial locations and with ``exemplars'', which are provided by the end user in the few-shot scenario.
In Section~\ref{sec:twostage}, 
we further introduce a two-stage training regime, 
in which the model is firstly pre-trained with self-supervision via masked image reconstruction~(MAE), followed by fine-tuning on the downstream counting task. To the best of our knowledge, this is the first work to show the effectiveness of self-supervised pre-training for generalised visual object counting.
Additionally, in Section~\ref{sec:mosaic}, we propose a novel and scalable {\em mosaic} pipeline for synthesizing training images, as a way to 
resolve the challenge of long-tailed distribution ({\em i.e.}~images with a large number of instances tend to be less frequent) in the existing object counting dataset.
In Section~\ref{sec:ttnorm}, we will introduce our test-time normalisation method including test-time normalisation and test-time cropping.



\subsection{Architecture}
\label{sec:arch}
\begin{figure}[t]
  \centering
  \includegraphics[width=0.98\textwidth]{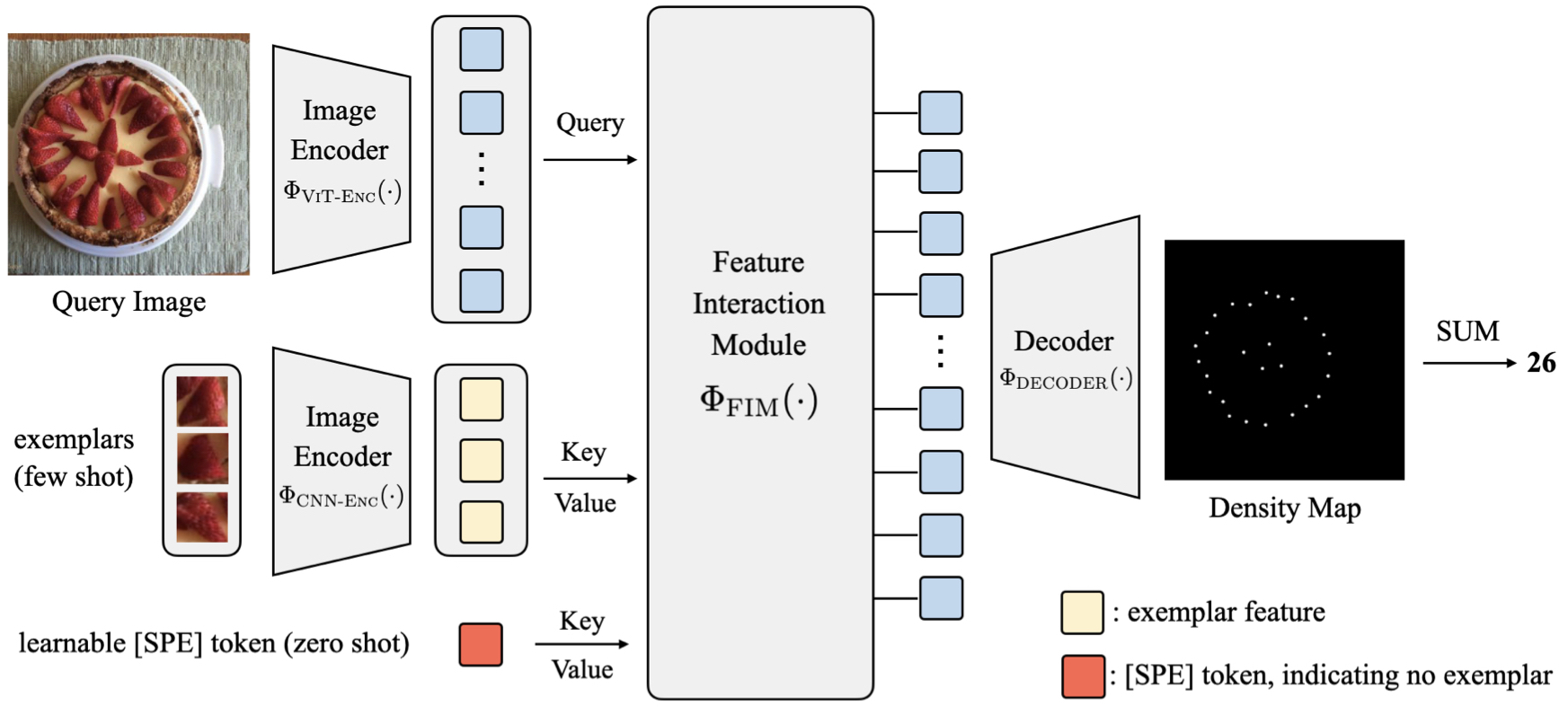} \\
  \caption{\textbf{Architecture detail for \textbf{CounTR}}. 
  The query image and exemplars are encoded by separate visual encoders.
  The image features are then fed into the feature interaction module as query vectors, and the exemplar features are fed as key and value vectors.  When there is no instance exemplar provided, 
  a learnable [\texttt{SPE}] token is used as key and value instead.
  The outputs are up-sampled in the decoder and finally, we get the corresponding density map. The object count can be obtained by summing the density map. 
  \textbf{Note that}, given the different exemplars with diversity,
  the model should ideally understand the invariance~(shape, color, scale, texture), for example, if the three given exemplars are all in the same color, the model should only count the objects of that color, otherwise, count all instances of the same semantic.
  }
 \label{fig:fewshot}
 \vspace{-10pt}
\end{figure}

Here, we introduce the proposed Counting TRansformer~(\textbf{CounTR}),
as shown in Figure~\ref{fig:fewshot}.
In specific, the input image~($\mathcal{X}_i$), 
and user-provided ``exemplars''~($\mathcal{S}_i^k, \forall k \in \{0,1, 2, 3\}$) are fed as input and mapped to a density heatmap, 
where the object count can be obtained by simply summing over it:
\begin{align}
    y_i =  \Phi_{\textsc{decoder}}( \Phi_{\textsc{FIM}}(\Phi_{\textsc{ViT-Enc}} (\mathcal{X}_i), \Phi_{\textsc{CNN-Enc}}(\mathcal{S}_i^k))),
    \hspace{3pt} \forall k \in \{0,1,...,K\}
\end{align}
In the following sections, we will detail the three building components, 
namely, visual encoder~($\Phi_{\textsc{ViT-Enc}}(\cdot)$ 
and $\Phi_{\textsc{CNN-Enc}}(\cdot)$), 
feature interaction module~({\em i.e.}~FIM, $\Phi_{\textsc{FIM}}(\cdot)$), 
and visual decoder~($\Phi_{\textsc{decoder}}(\cdot)$).


\subsubsection{Visual Encoder}

The visual encoder is composed of two components, 
serving for two purposes:
{\em first}, an encoder based on Vision Transformer~(ViT)~\cite{Dosovitskiy21} for processing the input image that maps it into a high-dimensional feature map;
{\em second}, compute the visual features for the ``exemplars'', 
if there is any.
Specifically, as for ViT, 
the input image is broken into patches with a size of $16 \times 16$ pixels and projected to tokens by a shared MLP.
To indicate the order of each token in the sequence, 
positional encoding is added, ending up with $M$ `tokens'.
They are further passed through a series of transformer encoder layers,
in our model, $12$ layers are used. 
We do not include the [CLS] token in the sequence, 
and the output from the ViT encoder is a sequence of $D$-dim vectors : 
\begin{align}
    \mathcal{F}_{\textsc{ViT}} = \Phi_{\textsc{ViT-Enc}}(\mathcal{X}_i) \in \mathbb{R}^{M \times D}
\end{align}
for more details, we refer the readers to the original ViT paper.

For few-shot counting, 
we use the exemplar encoder to extract the visual representation.
It exploits a lightweight ConvNet architecture~(4 convolutional layers, followed by a global average pooling), that maps the given exemplars~(resized to the same resolution) into vectors, 
\begin{align}
    \mathcal{F}_{\textsc{CNN}} = \Phi_{\textsc{CNN-enc}}(\mathcal{S}_i^k) \in \mathbb{R}^{K \times D}
\end{align}
Note that, under the zero-shot scenario with no exemplar given, we adopt a learnable [\texttt{SPE}] token as the substitute to provide cues for the model.

\subsubsection{Feature Interaction Module}
Here, we introduce the proposed feature interaction module~(FIM), 
for fusing information from both encoders.
Specifically, 
the FIM is constructed with a series of standard transformer decoder layers, where the image features act as the $\texttt{Query}$,
and two different linear projections~(by MLPs) of the exemplar features~(or learnable special token), 
are treated as the $\texttt{Value}$ and $\texttt{Key}$. 
With such design, the output from FIM remains the same dimensions as image features~($\mathcal{F}_{\textsc{ViT}}$), 
throughout the interaction procedure:
\begin{align}
    \mathcal{F}_{\textsc{FIM}} = \Phi_{\textsc{FIM}}(\mathcal{F}_{\textsc{VIT}},\text{ } W^{\texttt{k}} \cdot \mathcal{F}_{\textsc{CNN}}, \text{ }
    W^{\texttt{v}} \cdot \mathcal{F}_{\textsc{CNN}}) \in \mathbb{R}^{M \times D}
\end{align}
Conceptually, such transformer architecture perfectly reflects the self-similarity prior to the counting problem, 
as observed by Lu {\em et al.}~\cite{Lu18}.
In particular, the self-attention mechanisms in transformer decoder enables to measure of the self-similarity between regions of the input image, while the cross-attention between \texttt{Query} and \texttt{Value} allows to compare image regions with the \textbf{arbitrary} given shots, incorporating users' input for more customised specification on the objects of interest, or simply learn to ignore the ConvNet branch while encountering the learnable [\texttt{SPE}] token.

\subsubsection{Decoder}
At this stage, 
the outputs from the feature interaction module are further reshaped back to 2D feature maps and restored to the original resolution as the input image.
We adopt a progressive up-sampling design,
where the vector sequence is first reshaped to a dense feature map
and then processed by a ConvNet-based decoder.
Specifically, we use 4 up-sampling blocks, each of which consists of a convolution layer and a $2\times$ bilinear interpolation.
After the last up-sampling, we adopt a linear layer as the density regressor,
which outputs a one-channel density heatmap:
\begin{align}
    y_i = \Phi_{\textsc{decoder}}(\mathcal{F}_{\textsc{FIM}}) \in \mathbb{R}^{H \times W \times 1}
\end{align}

\subsection{Two-stage Training Scheme}
\label{sec:twostage}
In images, the visual signals are usually highly redundant, 
{\em e.g.}~pixels within local regions are spatially coherent.
Such prior is even more obvious in the counting problem, 
as the objects often tend to appear multiple times in a similar form.
Based on such observation,
we here consider exploiting the self-supervised learning to pre-train the visual encoder~($\Phi_{\textsc{ViT-Enc}}(\cdot)$). Specifically, we adopt the recent idea from Masked Autoencoders~(MAE), to train the model by image reconstruction with only partial observations.


\paragraph{Self-supervised Pre-training with MAE.}
In detail, we first divide the image into regular non-overlapping patches,
and only sample a subset of the patches (50\% in our case) as input to the ViT encoders. 
The computed features are further passed through a lightweight decoder, 
consisting of several transformer decoder layers,
where the combination of learnable mask tokens and positional encoding is used as \texttt{Query} to reconstruct the input image from only observed patches. The training loss is simply defined as the Mean Squared Error (MSE) between the reconstructed image and the input image in pixel space. \\[-0.6cm] 


\paragraph{Supervised Fine-tuning.}
After the pre-training, we initialise the image encoder with the weights of the pre-trained ViT, 
and fine-tune our proposed architecture on generalised object counting.
In detail,
our model takes the original image $\mathcal{X}_i$ and $K$ exemplars $\mathcal{S}_i = \{b_i\}^K$ from $\mathcal{D}_{\text{train}}$ as input and outputs the density map $\hat{y_i} \in \mathbb{R}^{H \times W \times 1}$ corresponding to the original image $\mathcal{X}_i$.
The statistical number of salient objects in the image $C_i \in \mathbb{R}$ can be obtained by summing the discrete density map $\hat{y_i}$.
We use the mean square error per pixel to evaluate the difference between the predicted density map $\hat{y_i}$ and the ground truth density map $y_i$.  The ground truth density maps are generated based on the dot annotations:
$\mathcal{L}(\hat{y_i}, y_i) = \frac{1}{HW} \sum || y_i - \hat{y_i}||_{2}^2$.


\subsection{Scalable Mosaicing}
\label{sec:mosaic}

\begin{figure}[t]
  \centering
    \subfigure[Type A: using four images.]{
    \begin{minipage}[t]{0.49\linewidth}
    \centering
    \includegraphics[width=0.95\textwidth]{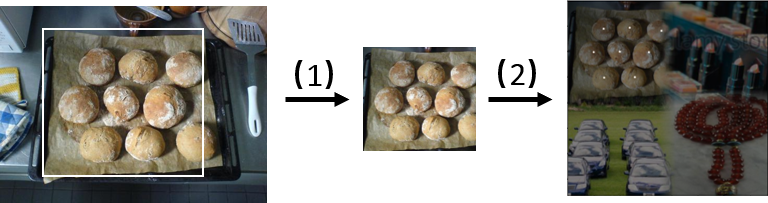}
    \end{minipage}%
    }%
    \subfigure[Type B: using one image.]{
    \begin{minipage}[t]{0.49\linewidth}
    \centering
    \includegraphics[width=0.95\textwidth]{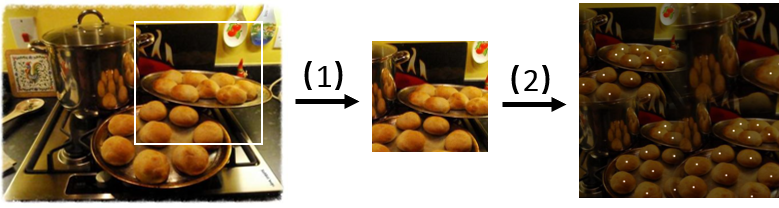}
    \end{minipage}%
    }%
    \vspace{-0.1cm}
    \caption{\textbf{The mosaic pipeline for synthesizing training images.} (1) stands for crop and scale, and (2) stands for collage and blending. In the following section, we will combine crop, scale, and collage as the collage stage. Type A uses four different images to improve background diversity and Type B uses only one image to increase the number of objects contained in an image. White highlights are the dot annotation density map after Gaussian filtering for visualization.}
  \label{fig:mosaic}
  \vspace{-0.5cm}
\end{figure}

In this section, we introduce a scalable {\em mosaic} pipeline for synthesizing training images, in order to tackle the long-tailed problem ({\em i.e.}~very few images contain a large number of instances) in existing counting datasets.
We observe that existing datasets for generalised object counting are highly biased towards a small number of objects.
For example, 
in the FSC-147 dataset, only 6 out of 3659 images in the train set contain more than 1000 objects. This is potentially due to the costly procedure for providing manual annotation. In the following, we elaborate on the two steps of the proposed mosaic training data generation,
namely, collage and blending (as shown in Figure~\ref{fig:mosaic}).
Note that, 
we also notice one concurrent work~\cite{Hobley22} uses a similar idea.\\[-0.6cm] 

\paragraph{Collage.}
Here, we first crop a random-sized square area from the image and scale it to a uniform size, {\em e.g.}~a quarter of the size of the original image.  
After repeating the region cropping multiple times, 
we collage the cropped regions together and update the corresponding density map.
It comes in two different forms: using only one image or four different images.
If we only use one image, we can increase the number of objects contained in the image, which helps a lot with tackling the long-tail problem.  If we use four different images, we can significantly improve the training images' background diversity and enhance the model's ability to distinguish between different classes of objects.  To fully use these two advantages, we make the following settings.
If the number of objects contained in the image is more than a threshold, we use the same image to collage; if not, we use four different images. Note that if four different images are used, we could only use the few-shot setting for inference, otherwise the model will not know which object to count. 
If we use the same image, 
the mosaiced image can be used to train the few-shot setting and zero-shot setting. \\[-0.8cm] 

\paragraph{Blending.}
Simply cropping and collaging do not synthesize perfect images, 
as there remain sharp artifacts between the boundaries. 
To resolve these artifacts, we exploit blending at the junction of the images.
In practise, 
we crop the image with a slightly larger size than a quarter of the original image size, such that we can leave a particular space at the border for $\alpha$-channel blending. We use a random $\alpha$-channel border width, 
which makes the image's composition more realistic.
\textbf{Note that}, we only blend the original image instead of the density map, 
to maintain the form of dot annotation (only 0 and 1).  Since there are few objects inside the blending border and the mosaic using one image is only applied to images with a very large number of objects, the error caused by blending is almost negligible.

\begin{figure}[t]
  \centering
    \subfigure[Test-time Normalisation.]{
    \begin{minipage}[t]{0.95\linewidth}
    \centering
    \includegraphics[width=0.98\textwidth]{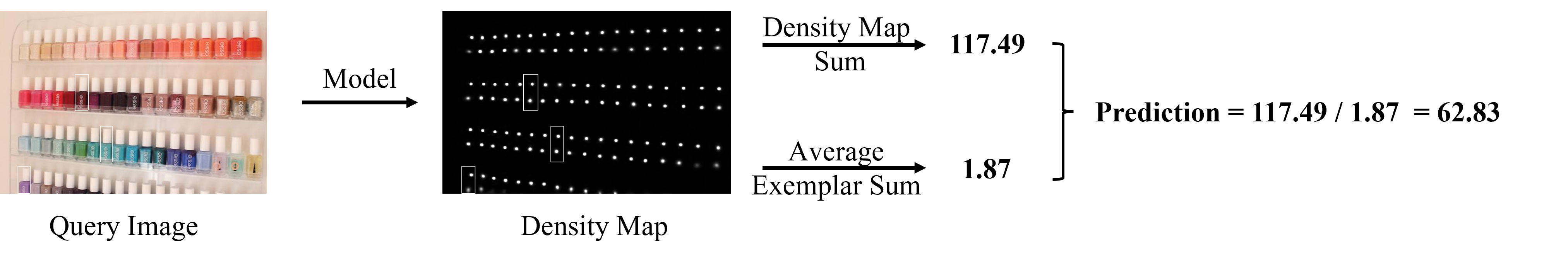}
    \end{minipage}%
    }%
    
    \subfigure[Test-time Cropping.]{
    \begin{minipage}[t]{0.95\linewidth}
    \centering
    \includegraphics[width=0.95\textwidth]{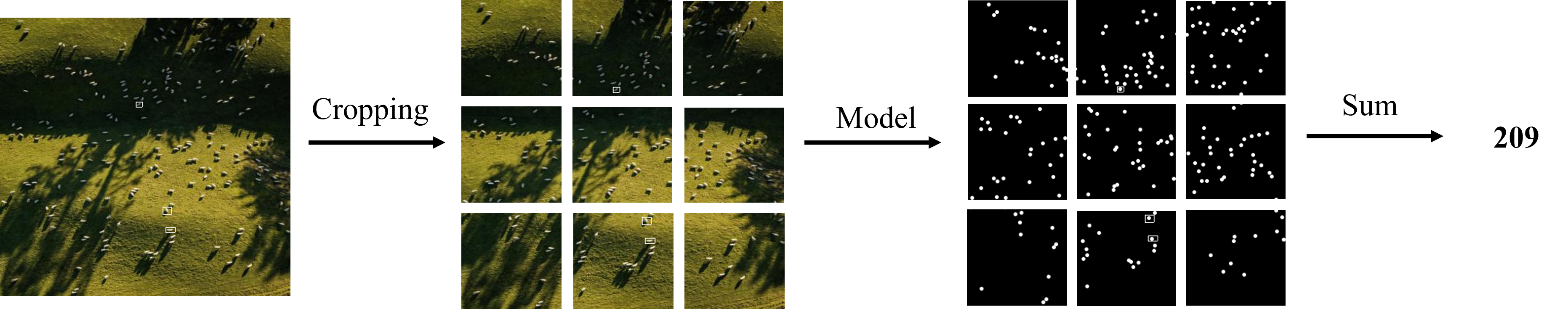}
    \end{minipage}%
    }%
    \caption{\textbf{The test-time normalisation process visualisation.}
    In test-time normalisation, if the average sum of the exemplar positions in the density map is over $1.8$, the sum of the density map will be divided by this average to become the final prediction.
    In test-time cropping, if at least one exemplar's side length is smaller than 10 pixels, the image will be cropped into 9 pieces and the model will process these 9 images separately. The final prediction will be the sum of the results of these 9 images.
    }
  \label{fig:tt}
  \vspace{-0.5cm}
\end{figure}

\subsection{Test-time Normalisation}
\label{sec:ttnorm}

For few-shot counting, 
we have introduced a test-time normalisation strategy to calibrate the output density map in the main text. 
Specifically, at inference time, we exploit the prior knowledge that 
the object count at the exemplar position should exactly be $1.0$,
any prediction deviation can thus be calibrated by dividing the density map by the current predicted count at the exemplar position.
We take this approach because due to the ambiguity of the bounding boxes, 
the model sometimes chooses the smallest self-similarity unit of an object to count, rather than the entire object, as shown in Figure~\ref{fig:tt}~(a).
Therefore, if the average sum of the density map area corresponding to the bounding boxes exceeds a threshold, such as 1.8, we will exploit this test-time normalisation approach.

Additionally, for images with tiny objects~(one exemplar with a side length shorter than 10 pixels), we adopt a sliding window prediction, 
by dividing the image equally into nine pieces and scaling them to their original size, to be individually processed by our model.  
The total number of objects is the sum of the individual count results of the nine images. 

\section{Experiments}
Here, we start by briefly introducing the 
few-shot counting benchmark, FSC-147 dataset, and the evaluation metrics;
In Section~\ref{sec:details}, 
we describe the implementation details of our model and the design of the inference stage; 
In Section~\ref{sec:sota}, we compare our model's performance with other counting models and demonstrate state-of-the-art performance on both zero-shot and few-shot settings; 
In Section~\ref{sec:ablation}, 
we conduct a series of ablation studies to demonstrate the effectiveness of the two-stage training and the image mosaicing;
In Section~\ref{sec:ae}, we give additional experiment results on Val-COCO, Test-COCO, and CARPK;
In Section~\ref{sec:qualitative},  we show the qualitative visualisation of CounTR's results on the FSC-147 dataset.\\[-0.5cm]

\subsection{Datasets and Metrics}
\label{sec:dataset}

\paragraph{Datasets.}
We experiment on FSC-147~\cite{Ranjan21}, 
which is a multi-class few-shot object counting dataset containing 6135 images.
Each image's number of counted objects varies widely, ranging from 7 to 3731, and the average is 56. The dataset also provides three randomly selected object instances annotated by bounding boxes as exemplars in each image. 
The training set has 89 object categories, while the validation and test sets both have 29 disjoint categories, making FSC-147 an open-set object counting dataset. \\[-0.6cm]

\paragraph{Metrics.}
We use two standard metrics to measure the performance of our model, 
namely, Mean Absolute Error (MAE) and Root Mean Squared Error (RMSE).  

\begin{equation}
    MAE = \frac{1}{N_I} \sum^{N_I}_{i=1}|C_i-C^{GT}_i|,
    \hspace{20pt}
    RMSE = \sqrt{\frac{1}{N_I} \sum^{N_I}_{i=1}(C_i-C^{GT}_i)^2}
\end{equation}

Here, $N_I$ is the total number of testing images, and $C_i$ and $C^{GT}_i$ are the predicted number and ground truth of the $i^{th}$ image.


\subsection{Implementation}
\label{sec:details}

\subsubsection{Training Details}
In this section, we aim to give the detail of our proposed two-stage training procedure,
that is, first pre-train the ViT encoder with MAE~\cite{he2022masked},
and then fine-tune the whole model on supervised object counting.\\[-0.6cm]

\paragraph{MAE Pre-training.}
As input, the image is of size $384 \times 384$,
which is first split into patches of size $16 \times 16$, 
and projected into $576$ vectors.
Our visual encoder uses 12 transformer encoder blocks with a hidden dimension of 768, 
and the number of heads in the multi-head self-attention layer is 12. 
The decoder uses 8 transformer layers with a hidden dimension of 512. 
As input for pre-training ViT with MAE, 
we randomly drop 50\% of the visual tokens, 
and task the model to reconstruct the masked patches with pixel-wise mean square error. During pre-training, 
we chose a batch size of 16 and trained on the FSC-147 for 300 epochs with a learning rate of $5 \times 10^{-6}$.\\[-0.6cm]

\paragraph{Fine-tuning stage.}
The feature interaction module uses 2 transformer decoder layers with a hidden dimension of 512. 
The ConvNet encoder exploits 4 convolutional layers and a global average pooling layer to extract exemplar features with 512 dimensions. 
The image decoder uses 4 up-sampling layers with a hidden dimension of 256. 
For optimisation, we minimise the mean square error between the model's prediction and the ground truth density map, which is generated with Gaussians centering each object. We scale the loss by a factor of 60, and randomly drop 20\% pixels, to alleviate the sample imbalance issue. 
We use AdamW as the optimiser.
Our model is trained on the FSC-147 training set with a learning rate of $1 \times 10^{-5}$ and a batch size of 8.
Our model is trained and tested on NVIDIA GeForce RTX 3090.

\subsubsection{Inference Details}
At inference time, we adopt sliding windows for images of different resolutions,
with the model processing a portion of an image with a fixed-size square window as used in training, and gradually moving forward with a stride of 128 pixels. 
The density map for overlapped regions is simply computed by averaging the predictions.\\[-0.5cm]

\vspace{-5pt}
\subsection{Comparison to state-of-the-art}
\label{sec:sota}

We evaluate the proposed CounTR model on the FSC-147 dataset and compare it against existing approaches. As shown in Table~\ref{tab:result},
CounTR has demonstrated new state-of-the-art on both zero-shot and few-shot counting, outperforming the previous methods significantly, especially on the results of the validation set.

\begin{table}[!htb]
\centering
\footnotesize
\setlength\tabcolsep{7pt}
 \begin{tabular}{cccccccc}
  \toprule
  \multirow{2}*{Methods} & \multirow{2}*{Year} & \multirow{2}*{Backbone} & \multirow{2}*{\# Shots}& \multicolumn{2}{c}{Val} & \multicolumn{2}{c}{Test} \\
  \cmidrule(lr){5-6}\cmidrule(lr){7-8}
  & & & &  MAE & RMSE & MAE & RMSE\\ 
  \midrule
    RepRPN-C~\cite{ranjan2022exemplar} & Arxiv2022 & ConvNets & 0 & 31.69 & 100.31 & 28.32 & 128.76\\
    RCC~\cite{Hobley22} & Arxiv2022 & Pre-trained ViT & 0 & 20.39 & 64.62 & 21.64 & 103.47 \\   
    \textbf{CounTR~(ours)} & 2022 & ViT & 0 & 18.07 & 71.84 & 14.71 & 106.87 \\
    \midrule
    FR~\cite{kang2019few} & ICCV2019 & ConvNets & 3 & 45.45 & 112.53 & 41.64 & 141.04 \\
    FSOD~\cite{fan2020few} & CVPR2020 & ConvNets & 3 & 36.36 & 115.00 & 32.53 & 140.65 \\
    P-GMN~\cite{Lu18} & ACCV2018 & ConvNets &3 & 60.56 & 137.78 & 62.69 & 159.67 \\
    GMN~\cite{Lu18} & ACCV2018 & ConvNets &3 & 29.66 & 89.81 & 26.52 & 124.57 \\
    MAML~\cite{finn2017model} & ICML2017 & ConvNets  & 3 & 25.54 & 79.44 & 24.90 & 112.68\\
    FamNet~\cite{Ranjan21} & CVPR2021 & ConvNets & 3 & 23.75 & 69.07 & 22.08 & 99.54 \\
    BMNet+~\cite{shi2022represent} & CVPR2022 & ConvNets & 3 & 15.74 & 58.53 & 14.62 & 91.83 \\
    \textbf{CounTR~(ours)} & 2022 & ViT & 3 & \textbf{13.13} & \textbf{49.83} & \textbf{11.95} & \textbf{91.23}\\
  \bottomrule
\end{tabular}
\captionof{table}{\textbf{Comparison with state-of-the-art on the FSC-147 dataset.}
P-GMN stands for Pre-trained GMN. RepRPN-C stands for RepRPN-Counter. RCC stands for reference-less class-agnostic counting with weak supervision.
}
\label{tab:result}
\vspace{-5pt}
\end{table}

\vspace{-0.3cm}
\subsection{Ablation Study}
\label{sec:ablation}
In this section, we have conducted thorough ablation studies to demonstrate the effectiveness of the proposed ideas, 
as shown in Table~\ref{tab:ablation},
we can make the following observations:
(1)~\textbf{Data augmentation:}
While comparing the Model-A, 
we include image-wise data augmentation in Model-B training, 
including Gaussian noise, Gaussian blur, 
horizontal flip, color jittering, and geometric transformation.
As indicated by the result, Model B slightly outperforms Model A on both validation and test set, suggesting that these augmentation methods can indeed be useful to the model to a certain extent, however, is limited. (2)~\textbf{Self-supervised pre-training:}
In Model-C, we introduce the self-supervised pre-training for warming up the ViT encoder. Compared with Model B which directly fine-tunes the ViT encoder~(pre-trained on ImageNet) on the FSC-147 training set, Model C has improved all results on both validation and test sets significantly. (3)~\textbf{Effectiveness of mosaic:}
With the help of the mosaic method, Model-D has shown further performance improvements, demonstrating its effectiveness for resolving the challenge from the long-tailed challenge, by introducing images with a large number of object instances, and object distractors from different semantic categories. (4)~\textbf{Test-time normalisation:}
In Model-E, we experiment with test-time normalisation for the few-shot counting scenario, where the output prediction is calibrated by the given exemplar shot. On both validation and test set, test-time normalisation has demonstrated significant performance boosts. (5)~\textbf{On shot number:}
In Model E, as the number of given shots increases, {\em e.g.}~1, 2, or 3, 
we observe no tiny difference in the final performance, 
showing the robustness of our proposed CounTR for visual object counting under {\em any} shots.\\[-0.4cm]

\begin{table}[!htb]
  \footnotesize
  \setlength\tabcolsep{3pt}
  \centering
     \begin{tabular}{cccccccccc}
      \toprule
      \multirow{2}*{Model} & \multirow{2}*{Augmentation} & \multirow{2}*{Selfsup} & 
      \multirow{2}*{Mosaic} & \multirow{2}*{TT-Norm.} & \multirow{2}*{\# Shots} &\multicolumn{2}{c}{Val} & \multicolumn{2}{c}{Test} \\
      \cmidrule(lr){7-8}\cmidrule(lr){9-10}
      & & & & & & MAE & RMSE & MAE & RMSE\\ 
      \midrule
       A0 & \XSolidBrush & \XSolidBrush & \XSolidBrush & \XSolidBrush& 0 & 24.84 & 86.33 & 21.06 & 130.04 \\
       A1 & \XSolidBrush & \XSolidBrush & \XSolidBrush & \XSolidBrush& 3 & 24.68 & 85.89 & 20.98 & 129.58 \\ 
       \midrule
       B0 & \Checkmark & \XSolidBrush & \XSolidBrush & \XSolidBrush& 0 & 23.80 & 81.53 & 21.14 & 131.27 \\
       B1 & \Checkmark & \XSolidBrush & \XSolidBrush & \XSolidBrush& 3 & 23.67 & 81.40 & 20.93 & 130.75 \\ 
       \midrule
       C0 & \Checkmark & \Checkmark & \XSolidBrush & \XSolidBrush& 0 & 18.30 & 72.21 & 16.20 & 114.30 \\
       C1 & \Checkmark & \Checkmark & \XSolidBrush & \XSolidBrush& 3 & 18.19 & 71.47 & 16.05 & 113.11 \\ 
       \midrule
       D0 & \Checkmark & \Checkmark & \Checkmark & \XSolidBrush & 0 & 18.07 & 71.84 & 14.71 & 106.87 \\
       D1 & \Checkmark & \Checkmark & \Checkmark & \XSolidBrush & 3 & 17.40 & 70.33 & 14.12 & 108.01 \\ 
       \midrule
       E1 & \Checkmark & \Checkmark & \Checkmark & \Checkmark & 1 & 13.15 & 49.72 & 12.06 & 90.01 \\
       E2 & \Checkmark & \Checkmark & \Checkmark & \Checkmark & 2 & 13.19 & 49.73 & 12.02 & 90.82 \\
       E3 & \Checkmark & \Checkmark & \Checkmark & \Checkmark & 3 & \textbf{13.13} & \textbf{49.83} & \textbf{11.95} & \textbf{91.23} \\ 
       E3~(no 7171.jpg) & \Checkmark & \Checkmark & \Checkmark & \Checkmark & 3 & 13.13 & 49.83 & 11.22 & 87.68 \\
       \bottomrule
    \end{tabular}
    \caption{\textbf{Ablation study}. 
    We observe that one image in the test set~(image id:7171) has significant annotation error~(see supp. material), result without it has also been reported.   
    \textbf{Selfsup}: refers to the proposed two-stage training regime.
    \textbf{TT-Norm}: denotes test-time normalisation}
    \label{tab:ablation}
    \vspace{-10pt}
\end{table}

\vspace{-0.5cm}

\subsection{Additional Experiments}
\label{sec:ae}
In this section, we further evaluate the model on several other datasets, 
{\em e.g.}, Val-COCO, Test-COCO, and CARPK. \\[-0.9cm]

\paragraph{Val-COCO and Test-COCO.}

Val-COCO and Test-COCO~\cite{Ranjan21} are FSC-147 subsets collected from COCO, 
and they are often used as evaluation benchmarks for detection-based object counting models. Here we compared our CounTR model with several counting models based on detection including Faster-RCNN~\cite{ren2015faster}, RetinaNet~\cite{lin2017focal}, and Mask-RCNN~\cite{he2017mask}.
As shown in Table~\ref{tab:COCOresult}, it can be easily found that our model still has a huge improvement even compared to the best-performing Mask-RCNN~\cite{he2017mask}, halving its error on both Val-COCO and Test-COCO.
We also compared our model with the few-shot counting sota method FamNet~\cite{Ranjan21}, and our model outperforms it with a large advance (15.16 MAE and 24.29 RMSE on Val-COCO and 11.87 MAE and 14.81 RMSE on Test-COCO), which demonstrates the superiority of our model.

\begin{table}[!htb]
\centering
\footnotesize
\setlength\tabcolsep{7pt}
 \begin{tabular}{cccccccc}
  \toprule
  \multirow{2}*{Methods} & \multirow{2}*{Year} & \multirow{2}*{Method} &  \multicolumn{2}{c}{Val-COCO} & \multicolumn{2}{c}{Test-COCO} \\
  \cmidrule(lr){4-5}\cmidrule(lr){6-7}
  & & &  MAE & RMSE & MAE & RMSE\\ 
  \midrule
    Faster-RCNN~\cite{ren2015faster} & NIPS2015 & Detection & 52.79 & 172.46 & 36.20 & 79.59 \\
    RetinaNet~\cite{lin2017focal} & ICCV2017 & Detection & 63.57 & 174.36 & 52.67 & 85.86 \\
    Mask-RCNN~\cite{he2017mask} & ICCV2017 & Detection  & 52.51 & 172.21 & 35.56 & 80.00 \\
    FamNet~\cite{Ranjan21} & CVPR2021 & Regression  & 39.82 & 108.13 & 22.76 & 45.92 \\
    \midrule
    \textbf{CounTR~(ours)} & 2022 & Regression & \textbf{24.66} & \textbf{83.84} & \textbf{10.89} & \textbf{31.11}\\
  \bottomrule
\end{tabular}
\captionof{table}{\textbf{Comparison with state-of-the-art on the FSC-147 subset.}
}
\vspace{-0.3cm}
\label{tab:COCOresult}
\end{table}

\paragraph{CARPK.}
CARPK~\cite{hsieh2017drone} is a class-specific car counting benchmark with $1448$ images of parking lots from a bird's view. 
We also fine-tuned our model on the CARPK train set and test on it with Non-Maximum Suppression (NMS).
We compared our CounTR model with several detection-based object counting models and regression-based few-shot counting models.
As shown in Table~\ref{tab:CARPKresult},
even compared with the existing class-specific counting models, 
{\em i.e.}, the models that can only count cars,
our CounTR still shows comparable performance.

\begin{table}[!htb]
\centering
\footnotesize
\setlength\tabcolsep{7pt}
 \begin{tabular}{cccccccc}
  \toprule
  \multirow{2}*{Methods} & \multirow{2}*{Year} & \multirow{2}*{Method} & \multirow{2}*{Type} &  \multicolumn{2}{c}{CARPK}\\
  \cmidrule(lr){5-6}
  & & & &  MAE & RMSE\\ 
  \midrule
    YOLO~\cite{redmon2016you} & CVPR2016 & Detection & Generic & 48.89 & 57.55 \\
    Faster-RCNN~\cite{ren2015faster} & NIPS2015 & Detection & Generic & 47.45 & 57.39  \\
    S-RPN~\cite{hsieh2017drone} & ICCV2017 & Detection & Generic & 24.32 & 37.62 \\
    RetinaNet~\cite{lin2017focal} & ICCV2017 & Detection & Generic & 16.62 & 22.30 \\
    LPN~\cite{hsieh2017drone} & ICCV2017 & Detection & Generic & 23.80 & 36.79 \\
    One Look~\cite{mundhenk2016large} & ECCV2016 & Detection & Specific & 59.46 & 66.84 \\
    IEP Count~\cite{stahl2018divide} & TIP2018 & Detection & Specific & 51.83 & - \\
    PDEM~\cite{goldman2019precise} & CVPR2019 & Detection & Specific & 6.77 & 8.52 \\
    \midrule
    GMN~\cite{Lu18} & CVPR2021 & Regression & Generic & 7.48 & 9.90 \\
    FamNet~\cite{Ranjan21} & CVPR2021 & Regression & Generic & 18.19 & 33.66 \\
    BMNet+~\cite{shi2022represent} & CVPR2022 & Regression & Generic & 5.76 & 7.83  \\
    \midrule
    CounTR~(ours) & 2022 & Regression & Generic & \textbf{5.75} & \textbf{7.45}\\
  \bottomrule
\end{tabular}
\captionof{table}{\textbf{Comparison with state-of-the-art on the CARPK dataset.}
}
\label{tab:CARPKresult}
\end{table}
\vspace{-0.5cm}

\subsection{Qualitative Results}
\label{sec:qualitative}

\begin{figure}[!htb]
  \centering
  \includegraphics[width=0.98\textwidth]{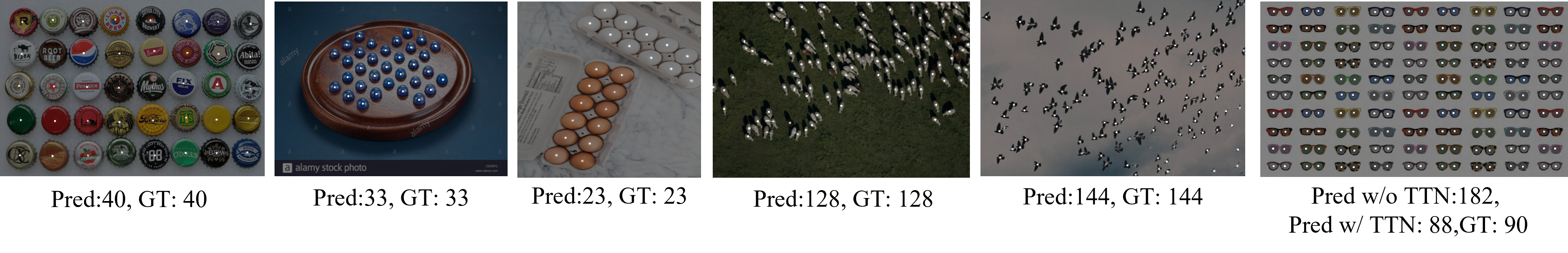} 
  \vspace{-5pt}
  \caption{\textbf{Qualitative results of CounTR on FSC-147.} 
  For visualisation purpose, 
  we have overlaid the predited density map on the original image. TTN stands for test-time augmentation.
  }
 \label{fig:good}
\end{figure}

\begin{figure}[!htb]
  \centering
  \includegraphics[width=0.99\textwidth]{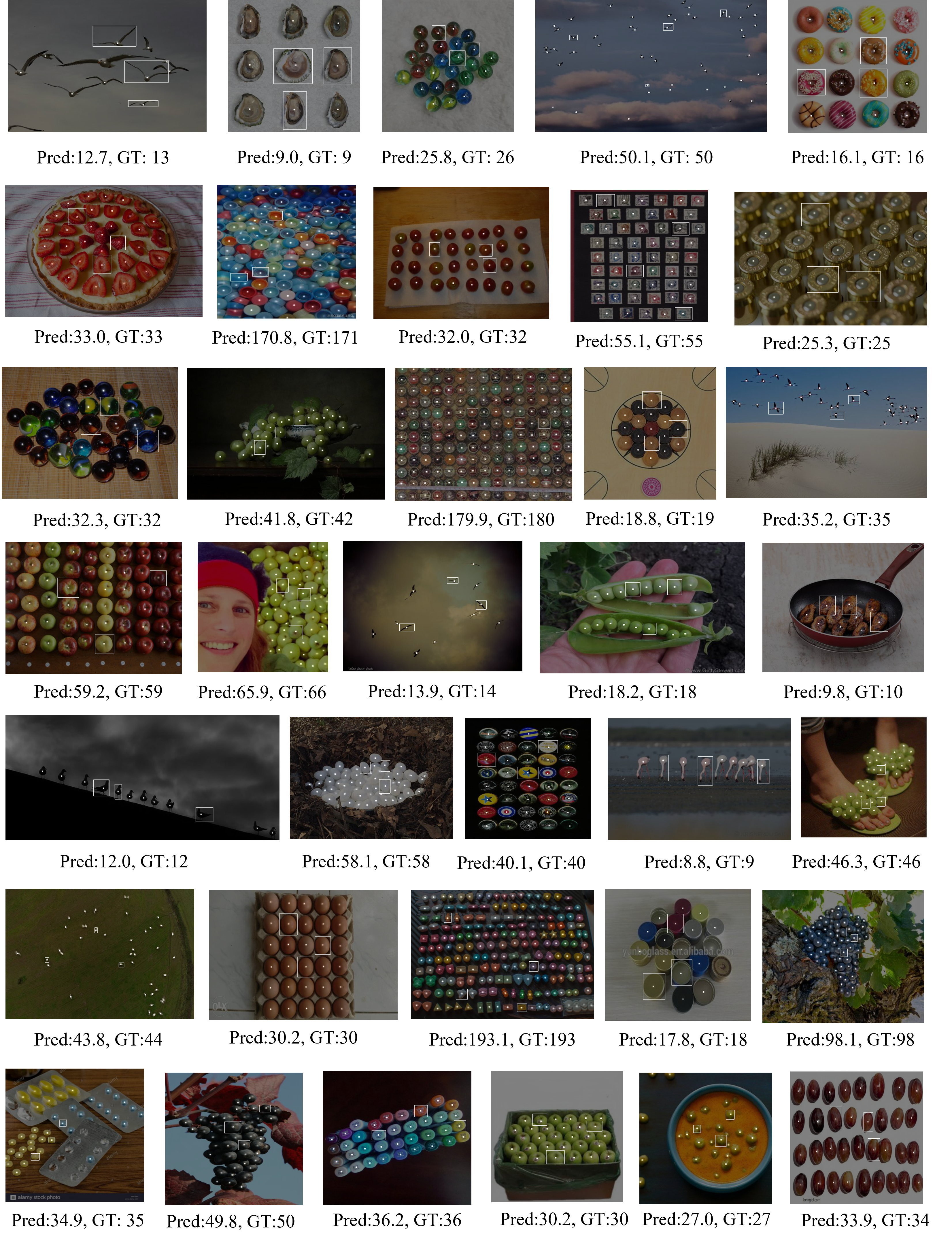} 
  \vspace{-5pt}
  \caption{\textbf{More qualitative results of CounTR on FSC-147.} 
  }
 \label{fig:good1}
\end{figure}

We show qualitative results from our few-shot counting setting in Figure~\ref{fig:good}.
As we can see from the first five images from FSC-147, 
our model can easily count the objects' numbers and locate their position.
The last image mistakenly chose the smallest self-similarity unit of spectacle lenses instead of sunglasses for counting due to the ambiguity of the bound boxes, which can be corrected by test-time normalisation.
For more qualitative visualisation, please refer to Figure~\ref{fig:good1}.

\section{Conclusion}
In this work, we aim at the generalised visual object counting problem of counting the number of objects from
{\em arbitrary} semantic categories using {\em arbitrary} number of “exemplars”. 
We propose a novel transformer-based architecture for it, termed as \textbf{CounTR}.
It is first pre-trained with self-supervised learning, and followed by supervised fine-tuning.
We also propose a simple, scalable pipeline for synthesizing training images that can explicitly force the model to make use of the given “exemplars”.
Our model achieves state-of-the-art performance on both zero-shot and few-shot settings.

\vspace{-0.2cm}
\paragraph{Acknowledgement} AZ is supported by EPSRC Programme Grant VisualAI EP/T028572/1, a Royal Society Research Professorship RP$\backslash$R1$\backslash$191132. We thank Xiaoman Zhang and Chaoyi Wu for proof-reading.

\bibliography{egbib.bib}

\clearpage  
\begin{appendix}

\section{Appendix}

In the supplementary material, 
we maily introduce the annotation error of 7171.jpg in the FSC-147 test set.


\subsection{On the annotation error of 7171.jpg}
\label{sec:7171}

We discover that the 7171.jpg in the FSC147 dataset has a significant annotation error. 
The annotated object count is inconsistent with the given exemplars, 
which tends to lead to a significant error during evaluation.
The ground truth annotation and exemplar annotation are shown in Figure~\ref{fig:279}.

\begin{figure}[!htb]
  \centering
    \subfigure[The exemplar annotation of 7171.jpg.]{
    \begin{minipage}[t]{0.4\linewidth}
    \centering
    \includegraphics[width=0.9\textwidth]{images/279bbox.png}
    \end{minipage}}%
    \subfigure[The ground truth annotation is 14.]{
    \begin{minipage}[t]{0.4\linewidth}
    \centering
    \includegraphics[width=0.9\textwidth]{images/279gt.png}
    \end{minipage}%
    }%
    \caption{The ground truth annotation and exemplar annotation of 7171.jpg, and we can easily figure out the inconsistency.}
  \label{fig:279}
\end{figure}

\end{appendix}
\clearpage

\end{document}